\pdfoutput=1
\documentclass[letterpaper]{article}
\usepackage[font=small]{caption}
\usepackage{proceed2e}
\usepackage[margin=1in]{geometry}

\usepackage{times}
\usepackage{graphicx}
\usepackage{subcaption}
\usepackage{colortbl}
\graphicspath{ {images/} }
\usepackage{amsmath}
\usepackage{times}
\usepackage{array}
\usepackage{multirow}
\usepackage{multicol}
\usepackage{tabularx}
\usepackage{booktabs}
\usepackage[table]{xcolor}
\usepackage[utf8]{inputenc}
\usepackage[english]{babel}
\usepackage[round]{natbib}
\usepackage{hyperref}
\usepackage[stable]{footmisc}
\title{PROFET: Construction and Inference of DBNs Based on Mathematical Models \thanks{To cite this paper, please use this full conference citation: 
	Hamda Ajmal, Michael Madden and Catherine Enright. PROFET: Construction and Inference of DBNs Based on Mathematical Models. In \textit{Proceedings of the 14th UAI Bayesian Modelling Applications Workshop (BMAW 2017)}, co-located with the 33rd Conference on Uncertainty in Artificial Intelligence (UAI 2017) Sydney, Australia, 2017.}}  
\author{Hamda Ajmal, Michael Madden and Catherine Enright\\School of Computer Science, National University of Ireland Galway\\
h.ajmal1@nuigalway.ie, michael.madden@nuigalway.ie, cathenright@gmail.com }
\begin{document}
\maketitle
\begin{abstract}
This paper presents, evaluates, and discusses a new software tool to automatically build Dynamic Bayesian Networks (DBNs) from ordinary differential equations (ODEs) entered by the user. The DBNs generated from ODE models can handle both data uncertainty and model uncertainty in a principled manner. The application, named PROFET, can be used for temporal data mining with noisy or missing variables. It enables automatic re-estimation of model parameters using temporal evidence in the form of data streams. For temporal inference, PROFET includes both standard fixed time step particle filtering and its extension, adaptive-time particle filtering algorithms. Adaptive-time particle filtering enables the DBN to automatically adapt its time step length to match the dynamics of the model. We demonstrate PROFET's functionality by using it to infer the model variables by estimating the model parameters of four benchmark ODE systems. From the generation of the DBN model to temporal inference, the entire process is automated and is delivered as an open-source platform-independent software application with a comprehensive user interface. PROFET is released under the Apache License 2.0. Its source code, executable and documentation are available at \url{http:://profet.it.nuigalway.ie}.
\end{abstract}
\section{Introduction}
Mathematical models of physical systems are widely available in many domains \citep{ottesen2004applied}. These embody existing expert knowledge and can be considered sufficient statistics of all prior experimentation in the domain. ODE models are generally available in mathematical, engineering and biological textbooks and research publications. 

Although these models have proven to be quite useful in many domains, they typically describe general population-level behaviors. For example, a large number of ODE models have been published to  investigate various dynamic aspects of cancer tumor growth and treatment \citep{eisen2013mathematical}. However, parameterization of these models is generic, typically done in a theoretical manner or based on laboratory data or literature-derived data. Therefore, they often fail to capture specific clinical scenarios. An individual patient's unique parameters can be considerably different from those of a population based model. 
To describe individuals, model parameters must be re-calibrated using observations of the individual. However, the observed data may be missing or noisy, or it could sparse or infrequent relative to the dynamics of the underlying system thus, making individualisation a challenging task. While ODE models are useful to understand the general treatments concepts, they are still not used for prescribing \emph{personalized} treatment regimes at the clinical level \citep{agur2014personalizing}.

Moreover, ODE models are completely deterministic with respect to their behavior, given a certain set of initial conditions \citep{vodovotz2009mechanistic}. Most real-world systems are rarely completely deterministic and they always contain some level of randomness or noise in them. The stochastic behavior of many real world systems indicates the need to account for stochasticity in ODE models. 

On the other hand, DBNs are well suited to handle uncertainty in a probabilistic fashion. However, discovering the structure of a DBN is a challenging task. While parameters are often learned from data, most DBN structures are constructed by hand, using knowledge elicited from domain experts \citep{chatterjee2010dbns}, which is a difficult and a time-consuming process \citep{lucas2004bayesian}.

Previous research work in our research group \citet{enright2013, enright2011,enright2013} has shown that mathematical models in the form of ODEs can be encapsulated into DBNs that incorporate a first order Euler solver, and can infer model values in future time slices. DBNs can reason efficiently with this powerful combination of domain knowledge and real-time data. They explicitly model measurement uncertainty and parameter uncertainty, allowing model parameters to be adjusted from initial approximate values to their correct values using real-time evidence. By doing this, the knowledge elicitation bottleneck is bypassed. The technique was previously applied to the problem of modeling glycaemia in patients in an Intensive Care Unit \citep{enright2010}, producing promising results.

This paper makes three key contributions. Firstly, we present a software application, PROFET, that automates the process of: (a) converting an ODE model to a DBN incorporating a first order Euler solver; and (b) inferring model values in future time slices using both standard fixed time step particle filtering \citep{gordon1993novel} and adaptive-time particle filtering \citep{enright2011} algorithms. Secondly, we evaluate PROFET by inferring model variables of four benchmark ODE models. Finally, we discuss various factors that impact the overall accuracy and performance of DBN inference.

Without PROFET, the process of generating a DBN from an ODE model is time-consuming and requires a detailed understanding of the underlying mathematics, as it involves the laborious and error-prone tasks of parsing the mathematical equations, creating the DBN manually, and writing code to execute the inference algorithms. PROFET automates the complete methodology and it can be used by researchers to carry out probabilistic inference from DBNs derived from mathematical models in any application domain.
\section{DBNs for Mathematical Models}
Mathematical models of different physical systems are generally available in the form of ODEs in domains such as mathematics, engineering, biology and bio-medicine. These mathematical models embody existing expert knowledge. However, as discussed earlier, they are usually idealizations as they attempt to describe a system's general dynamics, which can result in over-simplification and invite exception \citep{matos2001}. 
\subsection{Building the DBN Structure}
In order to \textit{individualize} a general mathematical model to a specific case, PROFET automatically maps it into a DBN. The DBN framework explicitly models noise as measurement and parameter uncertainty and then reduces the uncertainty over time by individualizing model parameters using temporal evidence. We automate the methodology we proposed in \citet{enright2013} to map a system of ODEs to a DBN which can also be expanded to reason effectively with noisy and temporal data. This methodology is described below with the help of an example.

Consider an Initial Value Problem: find {\small $X(t), Y(t),Z(t)$} such that {\small $X(t_0),Y(t_0),Z(t_0)$} are given and, 
{\small \begin{align*}
\label{eq:2}
\tag{2}
\begin{split}
\frac{dX}{dt} &= a.X(t) + Y(t).Z(t)
\\
\frac{dY}{dt} &= b(Y(t) - Z(t))
\\
\frac{dZ}{dt} &= c.Y(t) - Z(t) - X(t).Y(t)
\end{split}
\end{align*}} 
for all  $t > t_0$. This first order ODE system is known as the Lorenz model.  We will revisit this model in Section \ref{lor} for details. Here, $X$, $Y$ and $Z$ are the model variables and $a,b,c,d$ are the constant parameters. 
With PROFET the user simply enters the mathematical equations and the DBN structure is automatically built. The DBN approximates the values of model variables at times $t_1, t_2, t_3, ... $ using Euler's method, that is:
{\small \[X_{i+1} = X_{i} + h_{i}\frac{dX_i}{dt}\tag{3} \label{eq:3}\]} for $i = 0,1,...$, where $h_i = t_{i+1} - t_{i}$.
Thus, the rate of change of $X$ at step $i$ is 
{\small \[ \frac{dX_i}{dt} = \frac{X_{i+1} - X_{i}}{h_{i}} =:\Delta X_i  \tag{4} \label{eq:4}\]  }
and we can rewrite (\ref{eq:3}) as
{\small \[X_{i+1} = X_{i} + h_{i}\Delta X_{i} \tag{5} \label{eq:5} \text{for all $i = 0,1,...$}\] }
The same steps are followed to evaluate the values of model variables $Y$ and $Z$.
Figure \ref{fig:2} shows a graph of a DBN derived by PROFET from the Lorenz ODE model (\ref{eq:2}). The differentials are represented using an Euler approximation by mapping (\ref{eq:4}) and (\ref{eq:5}) directly to the six deterministic nodes, $\Delta X$, $X$,$\Delta Y$, $Y$, $\Delta Z$, $Z$ . The parents of the nodes $\Delta X, \Delta  Y, \Delta Z $ are set to be all the terms needed to evaluate them using (\ref{eq:4}), in the same time slice of the DBN. In each time slice, the values of $X_{i+1}$, $Y_{i+1}$ and $Z_{i+1}$ are evaluated using (\ref{eq:5}); hence the parents of nodes $X,Y,Z$ are set to be themselves and their corresponding  $\Delta$ nodes from the previous time slice. Extra inter-slice arcs on nodes $a,b,c,d$ allow parameters to be tuned to the evidence over time. Model parameters are modeled as continuous nodes. The probability distributions on these nodes depend on the individual ODE model being incorporated.

PROFET automates this methodology for any system of ODEs. It creates a sub-net for each equation in the model and adds dependencies between them, as dictated by the terms appearing in the right hand side of the equations. 
\subsection{Measurement Uncertainty of Observed Data}
The DBN provides a natural framework to deal with noisy data. The observed value of a variable that is to be approximated may contain an amount of measurement error. Following the algorithm we proposed in \citet{enright2013}, PROFET creates a separate node in the DBN for each observed value, which is dependent on its corresponding true value node. The observed variable nodes are continuous nodes; for each one, its mean ($\mu$) is the value of its parent node representing a true value and its standard deviation ($\sigma$) represents measurement uncertainty. Similarly, the actual inputs to a system may differ from the intended inputs, which are observed, and so a separate node for the intended (observed) input is added to the DBN. The true value node is conditionally dependent on its corresponding intended-value node. In the Lorenz model example, we assume that the values of $X$ are observed. It can be seen in Figure \ref{fig:2} that an extra evidence node (black) is present in the model to represent the relationship between observed value of the variable $X$ and its true value. 
\subsection{Model Parameter Re-Estimation}
\begin{figure}[h]
	\centering
\includegraphics[scale = 0.4]{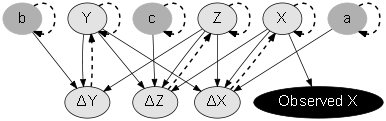}
	\caption{Lorenz ODE model transformed to a DBN incorporating first order Euler solver. }
	\label{fig:2}	
\end{figure}
The ODE model parameters are represented as continuous nodes in the DBN. They are allowed to vary in each time step. The value of a model parameter at each time step is conditionally dependent on its previous value, as shown by the dashed arcs in Figure \ref{fig:2}; they can therefore converge to values appropriate to the individual case over time, based on evidence from the temporal data streams.

A user begins by entering some initial estimate of the ODE model parameters distributions. The values of the distribution of the model parameters can often be obtained from published literature. If there are no published values or if the published population differs from the current population under study, a reasonable guess can be made or the ODE model parameters can be learned from the data from a group of observations using standard ODE model fitting techniques. 
As currently implemented, model parameters can have Uniform, Truncated Gaussian or Linear Gaussian distributions.  
\subsection{Evidence}
In PROFET, evidence can be defined as either continuous (which remains constant until a new value is reported) or instantaneous (where a value is specified for each moment in time). To account for noise, as was discussed  in Section 3.2, PROFET models observed nodes as continuous distributions whose mean ($\mu$) is the parent's node value, the true variable value and whose standard deviation ($\sigma$) represents measurement uncertainty. This  $\sigma$  can be configured by the user. Users can add continuous/instantaneous temporal data streams as evidence before running inference.
\subsection{Temporal Inference}
Two types of inference algorithms are implemented, fixed time step particle filtering, originally proposed by \citet{gordon1993novel} and adaptive-time particle filtering as we proposed in \citet{enright2013}. In fixed time step particle filtering, the step size is chosen by the user. To minimize the numerical error, the step size chosen must be sufficiently small. Results are to be reported at each step while filtering and prediction are carried out. But reducing the step size increases computational cost, so a trade-off must be made. This can be a challenge for stiff problems where very small step sizes are required, so inference quickly becomes inefficient. 

To overcome this limitation, PROFET also implements the adaptive-time particle filtering. The user specifies the intervals at which the results should be reported, and the adaptive-time inference algorithm automatically adopts a suitable time step that may be smaller than this. It aims to control the numeric error introduced at each time step, while minimizing run-times. To make this possible, the local error must be estimated which is done using delta nodes. This estimated error is compared to a prescribed tolerance. If the tolerance is met, the current step is accepted and a new step size is proposed for the next step, which may be bigger. If the tolerance is exceeded, the current step is rejected and a reduced step size is proposed. Both algorithms are described in detail in \citet{enright2013}.
\begin{figure}[h]
	\centering
	\includegraphics[width=0.5\textwidth]{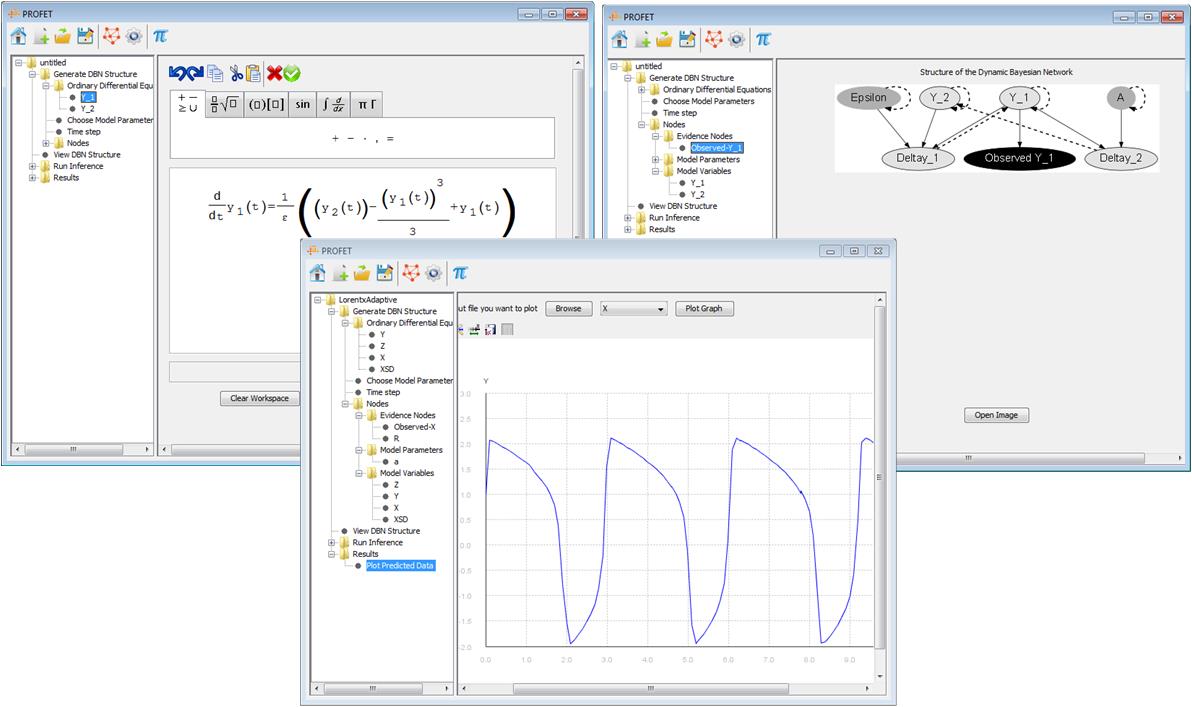}
	\caption{ Screenshots of PROFET GUI. Left: Drag and drop equation editor. Right: Graphical depiction of resultant DBN. Centre: Graph of inferred data.}
	\label{fig:3}	
\end{figure}
Results of the inference are saved and can be displayed in form of a graph inside PROFET, making it easier for users to comprehend and analyze. 
\section{Other Features of PROFET}
Figure \ref{fig:3} shows some screenshots of the PROFET GUI. Users can enter ODEs via a drag-and-drop equation editor which is based on DragMath \citep{sangwin2011}, modified with the authors' permission. The equations are automatically parsed into different types of DBN nodes and the DBN is constructed. The user has complete control over configuring model parameters.

PROFET uses GraphViz \citep{ellson2001} to draw DBNs in graphical form. The inferred data is summarized and can be displayed in form of chart, to facilitate analysis.

PROFET has been tested on Windows 7 and Ubuntu 14.04. Since it is developed in Java, it can run on any operating system that supports the Java Virtual Machine. The underlying algorithm for DBN structure construction and inference is written in Lisp. 
\section{Evaluation on Benchmark ODE Models}
In this section, we present four small to medium sized ODE models that we will use to benchmark inference and model parameter estimation in PROFET. Three out of these four models are biological systems. We chose them because \citet{dondelinger2013ode} applied their ODE model parameter inference methods to them. We evaluate the results of PROFET's ODE model variable inference and model parameter re-estimation on these ODE models; however our experimental setup is different from that of \citet{dondelinger2013ode}. The essence of their work is to infer the ODE model parameters from multiple noisy time series. Our aim is to \textit{individualize} the model parameters on a single time series data stream. Therefore, for each ODE model, our benchmark data is obtained using the R package deSolve. The values of the model parameters and the initial state of model variables to generate the benchmark data are taken from \citet{dondelinger2013ode}. Instead of adding noise to the data, we assume that the true values of the ODE model parameters are unknown and must be inferred from the population values. In order for PROFET to discover the correct model parameters, evidence (which in these cases are the true values taken from the benchmark solutions) is incorporated.

PROFET automatically converts the ODE models to DBNs. In all of our experiments, the model parameter nodes are modeled as continuous nodes with linear Gaussian distributions. The initial state model distribution can be viewed as the distribution of the population values. To simulate a situation where we do not know the true values of model parameters and/or initial state model variables of the data, we assume that we have a rough idea of the population parameters. In Section 6, we discuss how users can set the population mean ($\mu$) and the standard deviation ($\sigma_i$) of the model parameters for the initial state model. From one time step to the next, every model parameter is allowed to vary by setting $\mu$ equal to its value at the previous time step and  its standard deviation $\sigma_t$ on the transition model. Evidence for observed nodes is sampled from the benchmark data at different time points. To simulate a real world situation where evidence is sparse and infrequent, we deliberately sample evidence at sparse intervals. However, the evidence does not contain any noise.

In our experiments, we set the natural time step of the DBN equal to the step size used for ODE simulation to generate the benchmark data. For fixed time inference, we set the step size equal to the natural time step of the DBN. Particles are re-sampled and summarized at each step. For adaptive-time inference, our framework automatically selects an appropriate step size to match the dynamics of the system. For each experiment, we calculate Root Mean Square Error (RMSE) and Mean Absolute error (MAE) after an initial run-in period.

\begin{figure}[h]
	\begin{subfigure}[t]	{.22\textwidth}
		\centering
		\includegraphics[scale = 0.30]{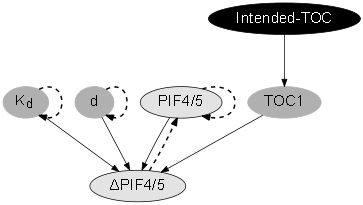}
		\caption{}
		\label{fig5:sfig1}
	\end{subfigure} 
	\hfil
	\begin{subfigure}[t]	{.22\textwidth}
		\centering
		\includegraphics[scale = 0.30]{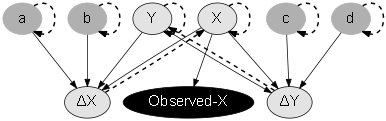}
		\caption{}
		\label{fig5:sfig2}		
	\end{subfigure}
%
	\caption{DBN structures for the ODE models (a) PIF4/5 model (b) Lotka-Volterra model 
	}
\end{figure}
\begin{table*}[]
	\centering
		\small
		
	\caption{Experimental Setup for Benchmark ODE models}
	\label{Setup}
	
	\setlength\tabcolsep{4pt}%
	\begin{tabular}{@{}lllllllllllllllll@{}}
		\toprule
		& \multicolumn{3}{l}{\textbf{PIF4/5 Model}} & \multicolumn{4}{l}{\textbf{Lotka Volterra Model}} & \multicolumn{6}{l}{\textbf{Signal Transduction Cascade Model}} & \multicolumn{3}{l}{\textbf{Lorenz Model}} \\

		\cmidrule(lr){2-4}
		\cmidrule(lr){5-8}
		\cmidrule(lr){9-14}
		\cmidrule(lr){15-17}
		\textbf{Natural time step }& \multicolumn{3}{c}{1} & \multicolumn{4}{c}{0.25} & \multicolumn{6}{c}{1} & \multicolumn{3}{c}{0.01} \\
		\midrule
		\textbf{Model Parameters} & s & $k_d$ & d & $\alpha$ & $\beta$ & $\gamma$ & $\delta$ & $k_1$ & $k_2$ & $k_3$ & $k_4$ & $k_m$ & $V$ & $a$ & $b$ & $c$ \\
		\cmidrule(lr){2-4}
		\cmidrule(lr){5-8}
		\cmidrule(lr){9-14}
		\cmidrule(lr){15-17}
		\multicolumn{1}{r}{$\mu$} & 0.7 & 0.50 & 1.3 & 2.2 & 1.15 & 4.2 & 0.8 & 0.09 & 0.3 & 0.08 & 0.3 & 0.1 & 0.04 & -8/3 & -10 & 28 \\
		\multicolumn{1}{r}{$\sigma_i$} & 0.1 & 0.05 & 0.1 & 0.1 & 0.1 & 0.1 & 0.1 & 0.01 & 0.1 & 0.01 & 0.1 & 0.1 & 0.01 & 0 & 0 & 1.12 \\
		\multicolumn{1}{r}{$\sigma_t$} & 0.01 & 0.01 & 0.01 & 0.05 & 0.05 & 0.05 & 0.05 & 0.001 & 0.01 & 0.001 & 0.01 & 0.01 & 0.001 & 0 & 0 & 0.1 \\
		\multicolumn{1}{r}{Benchmark} & 1 & 0.46 & 1 & 2 & 1 & 4 & 1 & 0.07 & 0.6 & 0.05 & 0.3 & 0.017 & 0.3 & $-8/3$ & -10 & 28 \\
		\midrule
		\textbf{Initial State} & \multicolumn{3}{l}{$PIF4/5$} & \multicolumn{2}{l}{$S$} & \multicolumn{2}{l}{$W$} & $S$ & $Sd$ & $R$ & $RS$ & \multicolumn{2}{l}{$Rpp$} & $X$ & $Y$ & $Z$ \\
		\cmidrule(lr){2-4}
		\cmidrule(lr){5-8}
		\cmidrule(lr){9-14}
		\cmidrule(lr){15-17}
		\multicolumn{1}{r}{Assumed} & \multicolumn{3}{l}{0.386} & \multicolumn{2}{l}{5} & \multicolumn{2}{l}{3} & 1 & 0 & 1 & 0 & \multicolumn{2}{l}{0} & 2 & 2 & 2 \\
		\multicolumn{1}{r}{Benchmark} & \multicolumn{3}{l}{0.386} & \multicolumn{2}{l}{5} & \multicolumn{2}{l}{3} & 1 & 0 & 1 & 0 & \multicolumn{2}{l}{0} & 1 & 1 & 1 \\
		\midrule
		\textbf{Inference }& \multicolumn{3}{l}{Fixed time step} & \multicolumn{4}{l}{Fixed time step} & \multicolumn{6}{l}{Fixed time step} & \multicolumn{3}{l}{Adaptive} \\
		\midrule
		
		\textbf{No. of samples} & \multicolumn{3}{l}{100,000} & \multicolumn{4}{l}{100,000} & \multicolumn{6}{l}{100,000} & \multicolumn{3}{l}{100,000} \\
		\midrule
		
		\textbf{Evidence times} & \multicolumn{3}{l}{{[}4,8,12,...,24{]}} & \multicolumn{4}{l}{{[}0.5,1.00,1.50,2.00{]}} & \multicolumn{6}{l}{{[}0,1,2,4,5,7,10,15,20,30,40,50,60,80,100{]}} & \multicolumn{3}{l}{[0.35,0.4,0.5...5]} \\
		\midrule
		\textbf{RMSE, MAE} & \multicolumn{3}{l}{0.070, 0.0353} & \multicolumn{4}{l}{0.287, 0.137} & \multicolumn{6}{l}{0.0085, 0.0053} & \multicolumn{3}{l}{ 0.250, 0.176} \\		
		\bottomrule		
	\end{tabular}	
\end{table*}

\subsection{The PIF4/5 Model}
The DBN created from the PIF4/5 ODE model is shown in Figure \ref{fig5:sfig1}. Initial values of model variables are set to the true values that are used to generate the benchmark solution. Benchmark data is generated by simulating the complete Locke 2-loop model. All other variables are discarded except $PIF4/5$ and $TOC1$. Evidence for the concentration of $TOC1$ is sampled from the benchmark data at each time step and for $PIF4/5$ at the time points shown in Table \ref{Setup}. As the evidence does not contain any noise, we set the standard deviation ($\sigma$) to a very small value, 0.01 to quantify the measurement uncertainty between the true value and the intended/observed values evidence nodes. The mean and the standard deviation of the likely values of the intended-$TOC1$ node are calculated from data from the benchmark solution.
We run the inference on the PIF4/5 model in PROFET from time $t=0$ to $t=24$ using standard fixed time step particle filtering using 100,000 particles. Complete details of experimental configuration and results are shown in Table \ref{Setup}.
\begin{figure}[h]
	\centering
	\begin{subfigure}[b]	{.22\textwidth}
		\centering
		\includegraphics[width=\textwidth]{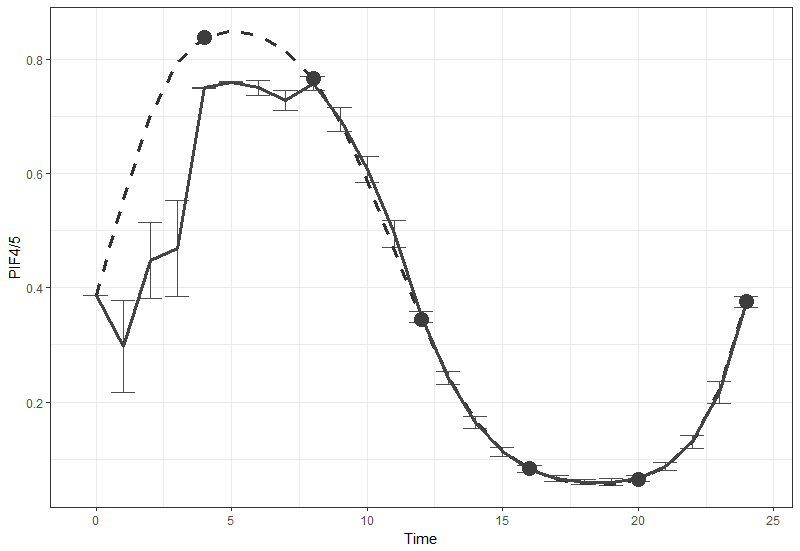}
		\subcaption{}
		\label{fig6:sfig1}
	\end{subfigure} 
	\hfil
	\begin{subfigure}[b]	{.22\textwidth}
		\centering
		\includegraphics[width=\textwidth]{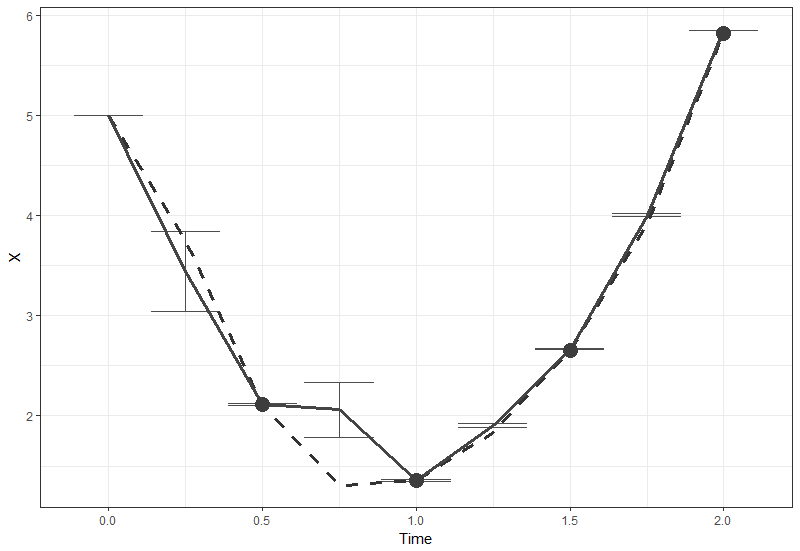}
		\subcaption{}
		\label{fig6:sfig2}		
	\end{subfigure}
	
	\begin{subfigure}[b]	{.22\textwidth}
		\centering
		\includegraphics[width=\textwidth]{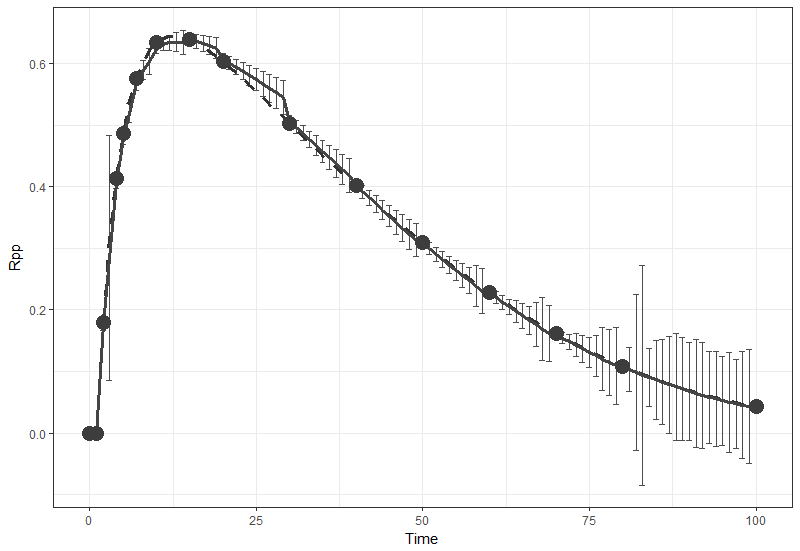}
		\caption{}
		\label{fig6:sfig3}			
	\end{subfigure}	
	\hfil
	\begin{subfigure}[b]{.22\textwidth}
		\centering
		\includegraphics[width=\textwidth]{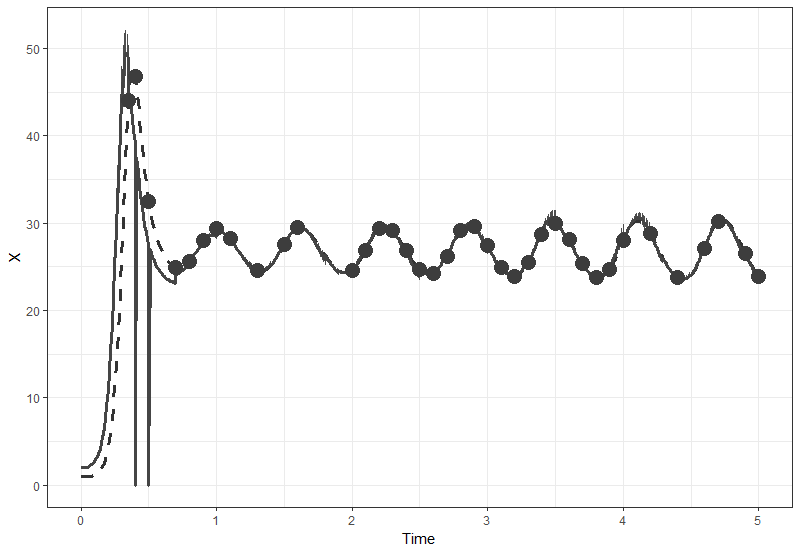}
		\caption{}
		\label{fig6:sfig4}				
	\end{subfigure}	
\caption{{\small Predicted values of model variables in the ODE system. Dashed lines represent the benchmark solution. Solid lines are the predicted trajectories. Error bars show one standard deviation. Gray dots are the points where evidence is received. (a) The PIF4/5 model (b) Lotka-Volterra Model (c) The signal transduction cascade model (d) The Lorenz model}}
\end{figure}
To validate the results, we compare the prediction accuracy of PROFET with the true benchmark solution. The graph in Figure \ref{fig6:sfig1} shows the comparison of benchmark solution and the predicted solution. It can be seen in the graph that the accuracy of the predicted values of $PIF4/5$ concentration begins to improve as evidence is incorporated into the system. Even though incorrect model parameters are chosen at the outset, they are re-estimated at every time step and eventually converge to their true values. The RMSE and MAE of the predicted values of $PIF4/5$ is calculated after first evidence is received at time $t=4$ and are shown in Table \ref{Setup}.
\subsection{The Lotka-Volterra Predator Prey Model}

We follow similar steps to those described in the previous section. The DBN constructed by PROFET is shown in Figure \ref{fig5:sfig2}. We run the inference on the Lotka-Volterra model in PROFET from time $t=0$ to $t=2$. Sparse evidence is provided from data of model variable $X$ sampled from the benchmark solution. $\sigma$ is set to 0.01.
In Figure \ref{fig6:sfig2}, we can see that as evidence is incorporated into the system, the prediction accuracy of PROFET improves. Even though incorrect values of model parameters were chosen at the start, the predicted results of PROFET are very close to those of benchmark solution. Full details of experimental configuration and results are shown in Table \ref{Setup}.

\subsection{The Signal Transduction Cascade}
Table \ref{Setup} again shows the experimental setup for the Signal Transduction Cascade model. We run the inference on the model in PROFET from time $t=0$ to $t=100$. Following \citet{dondelinger2013ode}, we sample evidence of $Rpp$ at more time points during the earlier part of the time series, where the dynamics tend to be faster. As before, $\sigma$ is set to 0.01.
In Figure \ref{fig6:sfig3}, we plot the data predicted by PROFET along with the benchmark solution. PROFET was able to predict the values of model variable $Rpp$ with high accuracy, even though incorrect model parameters are chosen at the outset. 
\subsection{The Lorenz Model}\label{lor}
In addition to the above three ODE models, we also evaluate the functionality of PROFET by using it to infer the values of a more challenging model, the Lorenz system of ODEs. The Lorenz model is a simplified mathematical model for atmospheric convection \citep{lorenz1963deterministic} developed by Edward Lorenz. It can have a chaotic solution under a certain set of model parameters and initial conditions. The model is a system of three ODEs (\ref{eq:2}).
The Lorenz model is known to be extremely sensitive to the initial conditions of the model variables.

In many real-world models, not all model variables have good initial values available, so these must be estimated from experience. This can lead to situations where the model must adjust rapidly over the first few time steps. To simulate such a scenario, we infer the model variables of the Lorenz model assuming that their initial values are unknown. 

We generate the benchmark data using ODE45 solver in R with values of model parameters as shown in Table \ref{Setup}, over the time interval $t=0$ to $t=5$. The resulting system is stiff: there are regions where the solution varies rapidly and standard numerical schemes often fail to yield a physically meaningful approximation to the solution unless extremely small step sizes are used. The time step used for ODE model simulation is 0.01. 

PROFET converts the Lorenz ODEs to a DBN, shown in Figure \ref{fig:2}. As this solution is stiff, we use the adaptive-time algorithm for inference. A tolerance of 0.1 is chosen to limit the numeric error introduced at each time step. Sparse evidence is sampled from the benchmark data.

It can be seen in the graph in Figure \ref{fig6:sfig4} that PROFET manages to find a good solution even though incorrect initial values of the model variables are used. It takes the system first few time steps to adjust to the true trajectory. There are some spurious spikes at time points 0.4 and 0.5 because the inferred value of $X$ is far from the evidence (benchmark), but the accuracy improves as more evidence is incorporated into the system. The RMSE and MAE are shown in Table \ref{Setup}. 

\section{Factors Affecting Performance}
We have demonstrated how PROFET can be utilized to run inference over ODE models whose parameters or the initial values of the model variables are unknown. However it must be noted that there are a few factors that impact the outcome of the inference. 

It is important that reasonable distribution means ($\mu$) and standard deviations ($\sigma_i$) are chosen for model parameters, as they can hugely impact the accuracy of inferred values. The  $\sigma_i$ of the initial state model parameter quantifies the amount of variation possible between the true model parameter value and the initially provided (potentially incorrect) value. The $\sigma_t$ on the transition model reflects how much the value of the model parameter is allowed to vary from its value at the previous time slice. A low transition $\sigma_t$ would make the model parameter at each time step close to its value in the previous time step, while a high $\sigma_t$ will allow bigger changes in the values of the model parameters. Thus, a large transition $\sigma_t$ is suitable for non-stationary problems. Similarly, the $\sigma$ on the observed nodes quantifies the measurement uncertainty of the observed data. If it is suspected that the evidence data is noisy, $\sigma$ should kept large enough to reflect that.

The choice of natural time step of the DBN and the step size used for inference is also very important, especially in stiff ODE models where they must be extremely small to capture the dynamics of the underlying system. E.g., if ODE model variables tend to change rapidly within one time unit, it is reasonable to select a smaller step size. Increasing the number of time steps increases the numerical accuracy of the solution. Of course, this increases the computational cost, but our adaptive-time inference algorithm seeks to mitigate this. In our experiments with the Lorenz model, we use a very small natural time step of the DBN (0.01) and the adaptive-time algorithm automatically adjusts the step size during inference.

The number of samples selected for particle filtering also plays an important role. As a rule of thumb, a higher number will improve accuracy, but at a price of increased computational cost. If a high variance is allowed for each node from the value at its previous node, it is reasonable to increase the number of samples, so that the sample space can be densely populated and there is a higher chance of filtering the values closer to the true value at each time slice. As would be expected, we also observe that the accuracy of the predicted results drops with an increase in the dimensionality of the model, keeping the number of samples fixed. As the number of model parameters whose values are to be estimated increases, the search space becomes sparse, and a larger number of samples are needed to find the best solution. 

The time required for inference increases linearly with the number of variables in the ODE and the number of particles. We do not anticipate scaling problems because real-world ODE models are usually formulated by domain experts and they do not generally involve a very large number of variables. 
\section{Related Work}
	\begin{table}[!htbp]
	\tiny
	\caption{Comparison of features of popular DBN software}
	\setlength\tabcolsep{3pt}%
	\begin{tabular}{*{5}{ p{0.06\textwidth} p{0.03\textwidth}p{0.1\textwidth}p{0.11\textwidth}p{0.11\textwidth}}}
		
		\toprule
		\textbf{Software} & \textbf{GUI} & \textbf{Structure Learning} & \textbf{Parameter Learning} & \textbf{Inference} \\ 
		\midrule
		GMTk & No & EM  & GEM & Frontier Algorithm\\ 
		SMILE \& GeNIe & Yes &	No & No & Yes (by unrolling DBN into a BN)\\
		Mocapy++ &	No & No & EM & Gibbs Sampling \\ 
		PNL & No &	Hill Climbing & EM &  1.5 Slice J-Tree algorithm \\
		libDAI & No & No & EM, MLE & Various \\ 
		BayesiaLab\textsuperscript{a} & Yes & SopLEQ, Taboo Search & Spiegelhalter and Lauritzen Parameterization Algorithm & J-Tree, Gibbs Sampling \\
		Hugin\textsuperscript{a} & 	Yes & No & EM & J-Tree 	\\
		Netica\textsuperscript{a} & Yes & No & Spiegelhalter and Lauritzen Parameterization Algorithm, EM and Gradient Descent &	Eliminiation J-Tree method \\ 
		BayesServer\textsuperscript{a} & Yes & PC Algorithm & EM &	Exact-Relevance Tree,
		Exact-Variable Elimination, Loopy Belief
		Propagation, Likelihood Sampling \\ 
		\midrule
		\textbf{PROFET} & Yes & Automatic mapping from ODEs to DBN incorporating first order Euler Solver &
		Model Parameters adjusted in real time by incorporating evidence&
		Particle Filtering, adaptive-time Particle Filtering \\ 
		\bottomrule
	\end{tabular}
	\textsuperscript{a} Commercial Software
	\label{table1}
\end{table}
Here, we present a review of somewhat related current software applications. In our survey, we have narrowed them down ten software applications that provide structure learning, parameter learning and inference on DBNs. Table \ref{table1} presents a feature comparison of these software applications.

There are also a few software packages for inference methods for ODE models. \textbf{BioBayes} \citep{vyshemirsky2008biobayes} is a software package that provides a framework for Bayesian parameter estimation of biochemical systems and evidential model ranking over models of biochemical systems defined using ODE models. For model parameter inference from experimental data, it implements a variant of Metropolis Hastings sampler \citep{hastings1970monte} and a population-based MCMC sampler \citep{jasra2007population}. Similarly, \textbf{ABC-SysBio} \citep{liepe2010abc} implements approximate Bayesian computation (ABC) for parameter inference and model selection in deterministic and stochastic ODE models. It combines ABC rejection sampler and ABC scheme based on Sequential Monte Carlo (ABC-SMC) \citep{toni2009approximate} for parameter inference. \textbf{GNU MCSim} \citep{bois2009gnu} is a numerical simulation and Bayesian statistical inference tool for algebraic or differential equation systems. \textbf{WinBUGS}  \citep{lunn2000winbugs} Differential Interface \textbf{(WBDiff)} \citep{lunn2004wbdiff} is an extension of WinBUGS that allows the numerical solution of any arbitrary systems of ODEs within WinBUGS models. The Runge-Kutta algorithm is used to solve the equations and Metropolis-Hastings \citep{hastings1970monte} samplers are used for sampling unknown inputs. \textbf{NIMROD} \citep{prague2013nimrod} facilitates the user to make approximate Bayesian inference in models with random effects based on ODEs. It is based on penalized maximum likelihood \citep{guedj2007maximum}.
The \textbf{Stan} programming language \citep{carpenter2016stan} can be used to fit the parameters of complex ODE models. It is a strongly-typed modeling language. Users can specify complex ODE models with minimal effort. It implements gradient-based MCMC algorithms for Bayesian inference, and gradient-based optimization for penalized MLE.

Bayesian Logic (\textbf{BLOG}) \citep{blog} is a probabilistic programming language with a declarative syntax. It is designed to describe probabilistic graphical models and then perform inference in those models. Assumed Parameter Filter \citep{APF} is an approximate inference algorithm that implements a hybrid of particle filtering for state variables and assumed density filtering for parameter variables in State Space Models (SSMs). As it is integrated into BLOG, it can be used within the framework. 

\textbf{LibBi} \citep{murray2013} is a software package for Bayesian inference specialized for SSMs designed for parallel computing on high-performance computer hardware. It implements SMC, particle Markov chain Monte Carlo (PMCMC) and SMC$^{2}$ for inference. 

In our survey, we did not find any software application that facilitates probabilistic ODE model variable inference using particle filtering by converting them to DBNs.

\section{Conclusions}
We have presented a user-friendly Java-based software application, called PROFET, that automatically converts first order ODE models to DBNs and performs temporal inference on them. The parameters of the DBN model are individualized as real time evidence is incorporated into the system. The software can be used by researchers in various domains interested in individualizing the general mathematical ODE models which are based on population level behavior.
We have evaluated PROFET by using it to infer model variable values of four benchmark ODE systems. PROFET can predict data with high accuracy and can deal with noisy, missing, sparse or infrequent evidence, incorrect model parameters and/or incorrect initial state values. We have also discussed some factors that affect the performance of the DBN inference. 

PROFET is free and open source. It is licensed under Apache License 2.0. The project website is \url{http://profet.it.nuigalway.ie/} and source code and executable are maintained on our GitHub account \url{https://github.com/HamdaBinteAjmal/PROFET} . No installation is required to run it. However the Java Run Time Engine must be installed on the user's machine. Detailed instructions and a comprehensive user manual, describing the software in a step-by-step approach through an example, are available on the website.

\renewcommand\bibsection{\subsubsection*{\refname}}
\small{
\bibliography{UAI}}
\end{document}